%% file: main.tex
\documentclass{article}
\pdfoutput=1

    \PassOptionsToPackage{numbers, compress}{natbib}

\usepackage[preprint]{neurips_2023}

\usepackage[utf8]{inputenc} 
\usepackage[T1]{fontenc}    
\usepackage{url}            
\usepackage{booktabs}       
\usepackage{amsfonts}       
\usepackage{nicefrac}       
\usepackage{microtype}      
\usepackage{xcolor}         
\usepackage{epsfig}
\usepackage{graphicx}
\usepackage{amsmath}
\usepackage{amssymb}
\usepackage{wrapfig}
\usepackage[pagebackref,breaklinks,colorlinks]{hyperref}

\title{CageViT: Convolutional Activation Guided Efficient Vision Transformer}

\author{Hao Zheng$^{1}$, Jinbao Wang$^{1}$, Xiantong Zhen$^{2}$, \\ \textbf{Hong Chen$^{3}$, Jingkuan Song$^{4}$, Feng Zheng$^{1\dagger}$}\\
$^1$Southern University of Science and Technology~ $^2$United Imaging Intelligence (UII)~ \\$^3$Huazhong Agricultural University \\$^4$University of Electronic Science and Technology of China
}

\begin{document}
\maketitle

\input{0_abs}
\input{1_intro}
\input{2_related}
\input{3_method}
\input{4_exp}
\input{5_conclusion}
\bibliographystyle{plain}
\bibliography{main}

\end{document}

%% file: 0_abs.tex
\begin{abstract}
Recently, Transformers have emerged as the go-to architecture for both vision and language modeling tasks, but their computational efficiency is limited by the length of the input sequence. To address this, several efficient variants of Transformers have been proposed to accelerate computation or reduce memory consumption while preserving performance. This paper presents an efficient vision Transformer, called CageViT, that is guided by convolutional activation to reduce computation. Our CageViT, unlike current Transformers, utilizes a new encoder to handle the rearranged tokens, bringing several technical contributions:
  1) Convolutional activation is used to pre-process the token after patchifying the image to select and rearrange the major tokens and minor tokens, which substantially reduces the computation cost through an additional fusion layer.
  2) Instead of using the class activation map of the convolutional model directly, we design a new weighted class activation to lower the model requirements.
  3) To facilitate communication between major tokens and fusion tokens, Gated Linear SRA is proposed to further integrate fusion tokens into the attention mechanism. We perform a comprehensive validation of CageViT on the image classification challenge.
  Experimental results demonstrate that the proposed CageViT outperforms the most recent state-of-the-art backbones by a large margin in terms of efficiency, while maintaining a comparable level of accuracy (\textit{e.g.} a moderate-sized 43.35M model trained solely on $224 \times 224$ ImageNet-1K can achieve Top-1 accuracy of 83.4\% accuracy).
\end{abstract}

\let\thefootnote\relax\footnotetext{$^\dagger$ Corresponding author.}

%% file: 1_intro.tex
\begin{figure*}[!ht]
\centering
\includegraphics[width=0.85\linewidth]{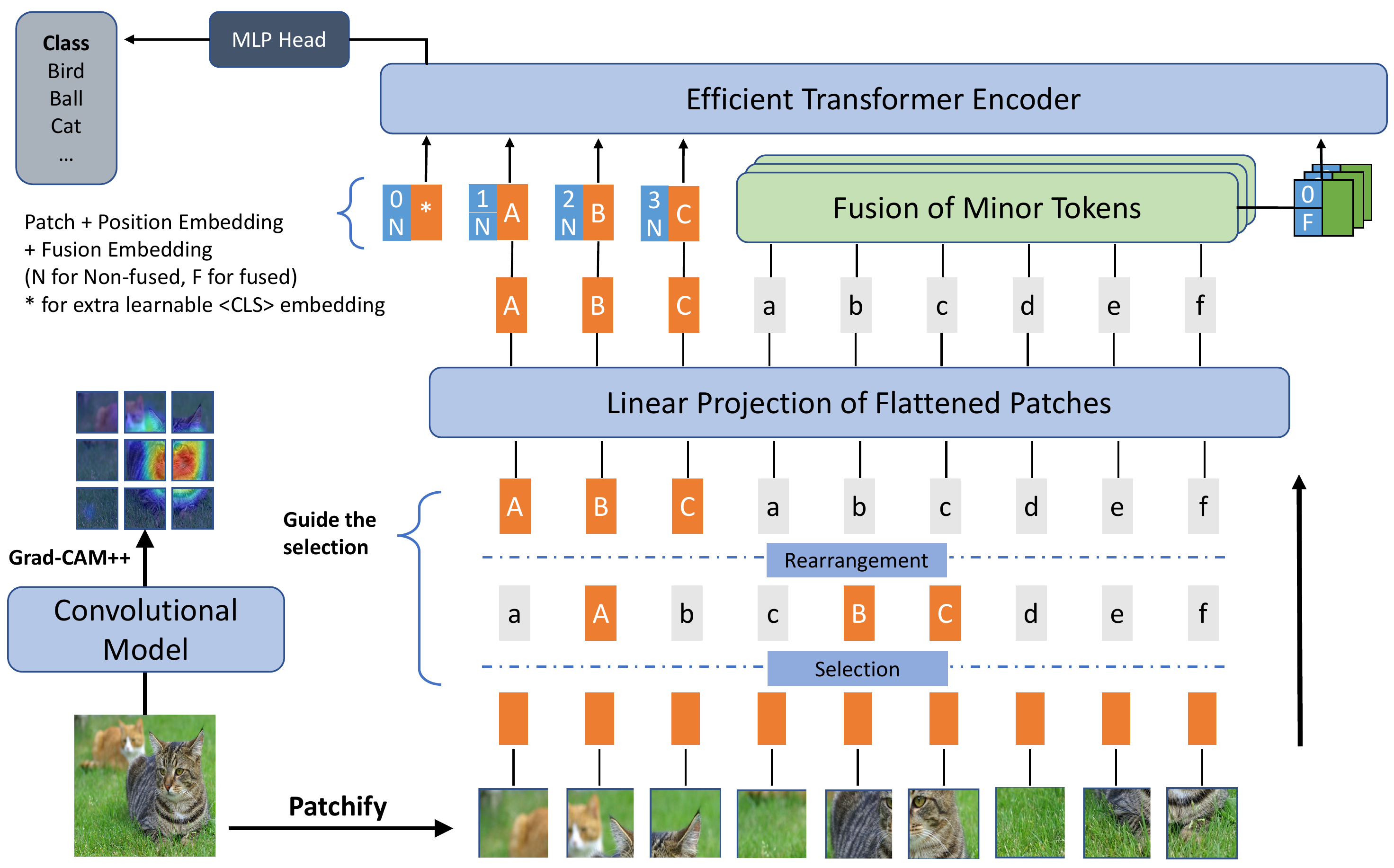}
   \caption{
   The overview of activation guided attention in our model. After partitioning the input image into patches of fixed size, the heat map produced by Grad-CAM++ via the auxiliary convolutional model $M_{conv}$ is utilized to choose the major tokens and the minor tokens. Then, for each patch, we perform a linear embedding. The minor tokens (grey tokens) are combined into several representation tokens ($N_f < N_m$) via a fusion network in order to decrease computation while preserving background knowledge. After incorporating positional embedding and fusion embedding into tokens, the output is fed into an efficient transformer encoder.
   }
\label{fig:method}
\vspace{-4mm}
\end{figure*}

\section{Introduction}
\label{introduction}

Transformer~\cite{vaswani2017attention} and its variations have achieved great success in both the visual and linguistic areas. The Transformer, for instance, is the backbone architecture of several advanced pre-trained language models in natural language processing, including BERT~\cite{devlin2018bert} and GPT~\cite{radford2019language}. In vision-related tasks, such as classification~\cite{dosovitskiy2020image}, object detection~\cite{carion2020end}, semantic segmentation~\cite{zheng2021rethinking}, pose estimation~\cite{lin2021end} and video captioning~\cite{zhou2018end}, Transformer also shows significant potential. Core to a Transformer model is the self-attention mechanism, which allows the Transformer to represent the contexts within an input sequence~\cite{parikh2016decomposable}. Due to the fact that self-attention computes the dot-product between each pair of input representations, its complexity is quadratic to the length of the input sequence~\cite{vaswani2017attention}. Therefore, it is challenging for standard Transformer models to effectively process lengthy input sequences~\cite{tay2020efficient}. In computer vision, however, many tasks~\cite{dosovitskiy2020image, wu2021cvt, wang2022pvt} demand high-resolution images that are transformed into lengthy series of image patches ahead of being processed using the Transformer model. Consequently, it is essential to construct an effective attention mechanism capable of modeling lengthy sequences.

Utilizing more efficient attention processes, many approaches have been proposed to accelerate the vanilla Transformer ~\cite{dai2019transformer, kitaev2020reformer, parmar2018image}. 
In one direction, the approaches lower the length of input tokens, such as ~\cite{wang2020linformer, tay2020synthesizer} by leveraging low-rank projections. These methods employ complex projection techniques to reduce computing effort while retaining feature representation. Due to the loss of fine-grained token-wise information, however, they frequently perform worse than full attention on tasks requiring fine-grained local knowledge, as evaluated by traditional language~\cite{tay2020long} and vision~\cite{zhang2021multi} benchmarks. In another direction, the approaches also attempt to sparsify the attention matrix using predefined patterns such as~\cite{tay2020sparse, parmar2018image, beltagy2020longformer, zaheer2020big}. These methods employ powerful inductive biases to enhance computing efficiency and model performance, but the self-attention layer's capabilities are severely constrained because each token can only attend to a subset of tokens.

To address the above limitations, we present the Convolutional Activation Guided Efficient Vision Transformer (CageViT), which speeds up the vanilla Transformer in both directions. Specifically, as shown in Figures~\ref{fig:method} and~\ref{fig:sra}, our CageViT is developed in three steps: 

1) Before feeding tokens into the linear projection layer, we execute an additional selection and rearrangement step based on the class activation map provided by Grad-CAM++. This process enables the network to identify in advance which patches are significant (major tokens) and which are not (minor tokens). 

2) After the linear projection of flattened patches, we combine minor tokens into multiple fusion tokens while retaining major tokens. Then, positional and fusion embeddings are added to tokens, and the results are sent to Transformer. This step can preserve the fine-grained information of major tokens, which dominate the classification process because determining the contours and textures is more significant in this task. 

3) We redesign the attention mechanism to improve communication between major tokens and fusion tokens, therefore facilitating the use of background information contained in fusion tokens to assist with classification. This step permits all major tokens to interact with minor tokens under the supervision of fusion tokens, \textit{i.e.}, they are gated by the fusion tokens, requiring the major token to be aware of the background information.

We thoroughly validate the effectiveness of the proposed CageViT using widely used benchmarks, such as image classification on ImageNet-1k. Compared to the most recent state-of-the-art Vision Transformers and CNNs, experimental results demonstrate the efficiency and generalization abilities of the proposed CageViT. Using a model size comparable to ResNet-18 (69.7\%) and PVT-Tiny (75.1\%), our CageViT-Small achieves Top-1 accuracy of 80.4\% on ImageNet-1k.

%% file: 2_related.tex
\section{Related Work}

Transformers~\cite{vaswani2017attention} is a sequence model that maintains the hidden activation of each time step and combines this data with the help of an attention operator~\cite{bahdanau2014neural}. Transformer will therefore represent the past with a tensor of past observations (depth $\times$ memory size $\times$ dimension). With this granular memory, Transformer has brought about a step-change in performance, particularly in machine translation~\cite{vaswani2017attention}, language modelling~\cite{dai2019transformer, shoeybi2019megatron}, video captioning~\cite{zhou2018end}, and a multitude of language understanding benchmarks~\cite{devlin2018bert, yang2019xlnet}. The computational expense of attending to every time step and the storage cost of retaining this enormous memory, \textit{i.e.} $O(n2)$ in both time and space complexity, is a disadvantage of storing everything. 
Multiple strategies~\cite{dai2019transformer, kitaev2020reformer, parmar2018image} have been proposed to reduce the computational cost of attention. 
These efficient transformers have a common name scheme: X-former, for instance, Reformer, \textit{e.g.} Reformer~\cite{kitaev2020reformer} and Performer~\cite{choromanski2020rethinking}. Based on previous research, efficient transformers may be loosely divided into three categories, namely fixed/factorized/random pattern-based approaches, low rank/kernels-based methods, and recurrence-based methods.

\textbf{Fixed/factorized/random pattern-based methods} perform by simply sparsifying the attention matrix. For instance, Image Transformer~\cite{dosovitskiy2020image}, which is inspired by CNNs, restricts the receptive field of self-attention to small local regions. This allows the model to process larger batch sizes while maintaining the likelihood of loss tractable. In addition, applying the notion of locality can be a desired inductive bias for image processing. 

Utilizing \textbf{low-rank} approximations of the self-attention matrix to enhance efficiency is a lately popular strategy. The central idea is to assume that the $N \cdot N$ matrix has a low-rank structure. A classic approach is shown by the Linformer~\cite{wang2020linformer}, which projects the length dimension of keys and values to a lower-dimensional representation ($N \rightarrow k$). Because the $N \cdot N$ matrix has been reduced to $N \cdot k$, it is evident that the low-rank technique alleviates the memory complexity issue of self-attention. 

\textbf{Recurrence-based methods} are an extension of the blockwise methods~\cite{qiu2019blockwise, parmar2018image} that link these chunked input blocks through recurrence. Transformer-XL~\cite{dai2019transformer} is the first to propose a system that connects multiple segments and blocks through segment-level recurrence. By using the last state as a memory state for the current segment, instead of computing the hidden states from scratch, the information is passed from state to state and not lost. This method effectively addresses the problem of semantic information loss and allows for longer-term dependencies.

%% file: 3_method.tex
\section{CageViT}

CageViT tries to approximate the full attention of the vanilla Transformer by aggregating all minor tokens into a single background token, while completely keeping the significant tokens. This section presents a preliminary overview of the multi-headed attention mechanism in the Transformer ~\cite{vaswani2017attention}. Then, we will detail how to discriminate major tokens from minor ones and fuse minor ones using the existing convolutional neural network. After that, we propose a new attention mechanism, called Gated Linear SRA, to further improve the model's efficiency. See Fig.~\ref{fig:method} for an illustration of our convolutional activation guided attention.

\subsection{Preliminaries and Notations}

Following ~\cite{vaswani2017attention}, the dot-product attention mechanism in vanilla Transformer can be defined as:
\begin{equation}
\label{equation Attn}
   \mathrm{Attention}(Q,K,V) = \mathrm{softmax}(\frac{QK^T}{\sqrt{d}})V,
\end{equation}
where $Q, K, V \in \mathbb{R}^{N \times d}$ are the query, key, and value embeddings, with $N$ as the sequence length and $d$ to be the hidden dimension.

Transformer's multi-headed attention (MHA) algorithm computes contextual representations for each token by paying attention to the entire input sequence across various representation subspaces. MHA is characterized as:
\begin{equation}
\label{equation MHA}
   \mathrm{MHA}(Q,K,V) = \mathrm{Concat}(\mathrm{head}_0,...,\mathrm{head}_{N_h})W^O,
\end{equation}

\begin{equation}
\label{equation head}
   \mathrm{head}_i = \mathrm{Attention}(Q_i, K_i, V_i)
\end{equation}
for the $i^{th}$ head, where $N_h$ is the number of head, $W^O \in \mathbb{R}^{Nd \times d}$ is the projection matrix to project $nd$ dimension into $d$ dimension.

\subsection{Convolutional Activation Guided Attention}
\label{CAGA}

In this section, we describe how to incorporate convolutional activations in the Transformer. As shown by Eq.~(\ref{equation Attn}), the inner product between $Q$ and $k$ has both time and space complexity of $O(N^2)$. One strategy to increase efficiency is to reduce the number of tokens, or $N$. In image classification tasks, humans are able to quickly distinguish between important, or ``major" tokens, and less important, or ``minor" tokens. Thus, by combining all minor tokens into a few tokens with all the background information before they enter the Transformer, we can greatly reduce computational complexity while still retaining all the important major tokens for accurate classification.

In this paper, we use class activation maps created by Grad-CAM++~\cite{chattopadhay2018grad} to identify major and minor tokens. Since the class label is necessary for Grad-CAM++ to produce the class activation map, which is not available during testing, we choose the top-$K$ labels based on the confidence of the final layer and calculate a weighted average using this confidence. The resulting salience map from the weighted average Grad-CAM++ is defined as:
\begin{equation}
\label{equation weighted}
   S_{ij} =  \sum_{k}^{K} ( \frac{z^k}{Z} \cdot L_{ij}^{c_k} ),
\end{equation}
where $z_k$ is the confidence of the last layer for class label $c_k$, $Z = \sum_{k}^{K} z_k $ is the regularization term, $L_{ij}$ is the value of salience map given by Grad-CAM++ at the $i^{th}$ row and $j^{th}$ column.

After obtaining the class activation map, the importance of each patch can be calculated by summing the activation values within the patch. In this paper, we choose a ratio of $\rho$ major tokens out of the total tokens. These tokens are then rearranged and fed into the linear projection layer, as shown in Fig.~\ref{fig:method}.

After passing through the linear projection layer, the selected minor tokens are fed into a multi-head fusion layer composed of multiple multi-layer MLPs to reduce the token dimension. The multi-head fusion layer is defined as follows:
\begin{equation}
\label{equation MHF}
   \mathrm{MHF}_i(m) = MLP(\mathrm{Concat}(m_0, m_1, m_2, ..., m_{N_m}),
\end{equation}
where for the $i^{th}$ fusion head, $N_m = \lfloor N_b \cdot (1-\rho) \rfloor$ is the number of minor tokens, $N_b$ is the total number of batches, and $m_i$ is the $i^{th}$ selected minor token.

\textbf{Extra model.} We add an additional convolutional model $M_{conv}$, specifically MixNet-S~\cite{tan2019mixconv}, to the Transformer, requiring extra compute and memory during training and testing. However, based on the results presented in Sec.\ref{impact of conv}, we found that we do not need a high-accuracy convolutional model to achieve good results. We can use a convolutional model with average performance (50\% accuracy on ImageNet) which is computationally efficient. Experiments involving the forwarding process of $M_{conv}$, necessitates the selection of a lightweight convolutional model. The performance of MixNet-S can be found in Tab.~\ref{table: main}.

\begin{figure*}
\begin{center}
\includegraphics[width=0.8\linewidth]{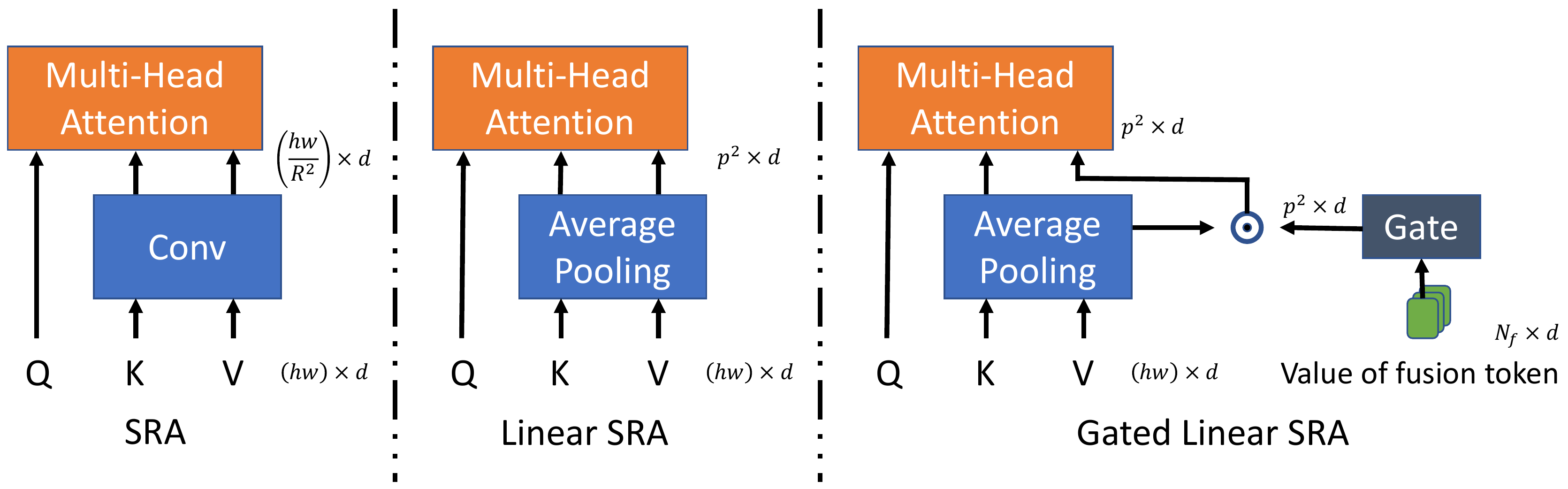}
\end{center}
   \caption{
   Comparison of SRA in PVTv1~\cite{wang2021pyramid}, Linear SRA in PVTv2~\cite{wang2022pvt}, and our proposed Gated Linear SRA, which can make it easier to use the contextual information stored in the fusion tokens to aid classification. All the fusion tokens are concatenated together and fed into the Gate layer.
   }
\vspace{-2mm}
\label{fig:sra}
\end{figure*}

\subsection{Gated Linear Spatial Reduction Attention}
Here we present Gated Linear Spatial Reduction Attention (Gated Linear SRA), a redesigned layer that better fits our proposed CAGA. We start by introducing the related layers:

\textbf{Spatial Reduction Attention (SRA).} SRA ~\cite{wang2021pyramid} is proposed to reduce attention operation's computational and memory costs for high-resolution feature maps. The key idea is to reshape the key and value from $(hw) \times d$ to $\frac{hw}{R^2} \times d$ in each encoder layer.
\begin{equation}
\label{equation SRA}
   \mathrm{SRA}(Q,K,V) = \mathrm{Concat}(\mathrm{head}_0,...,\mathrm{head}_N)W^O.
\end{equation}
Here each head is obtained by
\begin{equation}
\label{equation SRA head}
   \mathrm{head} = \mathrm{Attention}(Q, SR(K), SR(V)),
\end{equation}
where $SR(\cdot)$ is the operation for reducing the spatial dimension of $K$ and $V$, which can be written as:
\begin{equation}
\label{equation SR}
   \mathrm{SR}(x) = \mathrm{Norm}(\mathrm{Reshape}(x, R)W^S).
\end{equation}

Here, $R$ is the reduction ratio, $x \in \mathbb{R}^{hw \times d}$ denotes the input sequence. $Reshape$ is an operation to reshape the input sequence into $\frac{hw}{R^2} \times d$, which can be implemented with a convolution over the 2-d dimension of the feature map. The complexity of SRA is:
\begin{equation}
\label{equation Omega SRA}
   \Omega(SRA) = \frac{2h^2w^2d}{R^2} + hwd^2R^2.
\end{equation}

\textbf{Linear SRA.} Different from SRA, Linear SRA ~\cite{wang2022pvt} uses average pooling for spatial dimension reduction, which can reduce the dimension into a fixed size $(p^2 \times d)$ before attention operation. Thus, the complexity of Linear SRA can be reduced to a linear level:
\begin{equation}
\label{equation LSRA MHA}
   \Omega(Linear SRA) = 2hwp^2d,
\end{equation}
where $p$ is pooling size of Linear SRA, and is set to 7 in PVTv2 ~\cite{wang2022pvt}.

\textbf{Gated Linear SRA.} The primary concept behind our proposal is to incorporate the fusion token, introduced in Section~\ref{CAGA}, deeper into the attention calculation process. By merging all minor tokens into fusion tokens, the fusion token carries a wealth of environmental information that can better guide the attention. Also, since the locality information disrupted by the selection and rearrangement is still preserved in the environment representations, the introduced gating mechanism can facilitate the communication between major tokens and fusion tokens. Specifically, Gated Linear SRA can be formulated as follows:
\begin{equation}
\label{equation Ghead}
   \mathrm{head} = \mathrm{Attention}(Q, SR(K), SR(V) \odot \mathrm{Gate}(V^f)),
\end{equation}
where $\odot$ stands for element-wise multiplication, $V^f$ is the concatenated value of fusion tokens in each layer, and $SR(\cdot)$ is the same operation used in Linear SRA, \textit{i.e.} average pooling to a fixes size of $(p^2 \times d)$.
Note that the Gate module is a two-layer MLP that is used to map multiple fusion tokens into dimensions that can exchange information with the average-pooled value token.

Since the complexity of MLP and element-wise multiplication is linear, the complexity of Gated Linear SRA is equivalent to that of Linear SRA:
\begin{equation}
\label{equation GLSRA}
   \Omega(Gated Linear SRA) = \Omega(Linear SRA) = 2hwp^2d.
\end{equation}

\begin{table*}
\small
  \caption{Details of CageViT variants. We also report the total parameters adding the MixConv-s~\cite{tan2019mixconv} (4.10M) between brackets.}
  \centering
  \begin{tabular}{lcccccccccc}
    \toprule
    Model   & $L$  & $d$ & $D$ & $p$ & $N_h$ & $N_f$ & $K$ & $\rho$ (\%) & \#Params (M)   \\
    \midrule
    CageViT-T  & 8 & 768 & 1024 & 7 & 8 & 4 & 9 & 20 & 9.91 (14.01) \\
    CageViT-S  & 8 & 768 & 1024 & 7 & 12 & 8 & 9 & 20 & 13.53 (17.63) \\
    CageViT-B  & 12 & 768 & 2048 & 7 & 12 & 8 & 9 & 20 & 24.34 (28.44) \\
    CageViT-L  & 16 & 1024 & 2048 & 7 & 16 & 8 & 9 & 20 & 43.35 (47.45) \\
    \bottomrule
  \end{tabular}
\label{table: variants}
\end{table*}

\subsection{Model Variants}

In summary, the hyperparameters of our method are listed as follows:
\begin{itemize}
\itemsep0em
  \item $L$: the number of encoder layers;
  \item $d$: the size of hidden dimension in the encoder layer;
  \item $D$: the size of hidden dimension in MLP head layer;
  \item $p$: a constant value that is used to reduce the spatial dimension in Gated Linear SRA;
  \item $N_h$: the head number of the Efficient Self-Attention;
  \item $N_f$: the number of fusion heads in the multi-head fusion module in Sec.~\ref{CAGA};
  \item $K$: the number of class labels used to calculate the weighted average of activation in Eq.~(\ref{equation weighted});
  \item $\rho$: the ratio of major tokens selected by CAGA in Sec.~\ref{CAGA};
\end{itemize}

To provide instances for discussion, we describe a series of CageViT model variants with different scales, namely CageViT-T, CageViT-S, CageViT-B, CageViT-L (Tiny, Small, Base, and Large, respectively), in Tab.~\ref{table: variants}. As is said in Sec.~\ref{CAGA}, a light-weighted convolutional model is required for the calculation of Grad-CAM++, so MixConv-s~\cite{tan2019mixconv} is chosen in the experiments. We also report the total parameters by adding the MixConv-s (4.10M) together with our CageViT in Tab.~\ref{table: variants}.

%% file: 4_exp.tex
\begin{table*}[!ht]
\small
  \caption{Comparison with state-of-the-art backbones on ImageNet-1k benchmark. Throughput(images / s) is measured on a single V100 GPU, following~\cite{touvron2021training}. All models are trained and evaluated on $224 \times 224$ resolution. The parameters of our models are reported in two parts: CageViT parameters and total parameters (in parentheses), where total parameters take the extra convolutional model $M_{conv}$ into account.}
  \label{table: main}
  \centering
  \begin{tabular}{lccccc}
    \toprule
    Model     & \#Params (M)     & FLOPs (G)  & Throughput & Top-1 (\%) & Top-5 (\%) \\
    \midrule
    \multicolumn{6}{c}{ConvNet}                   \\
    \midrule
    ResNet-18 ~\cite{he2016deep}                    & 11.7 & 1.8  & 1852 & 69.7 & 89.1 \\
    MixNet-S ~\cite{tan2019mixconv}                 & 4.1 & 0.256 & -  & 75.8 & 92.8 \\
    MixNet-M ~\cite{tan2019mixconv}                 & 5.0 & 0.36 & -  & 77.0 & 93.3 \\
    ResNet-50 ~\cite{he2016deep}                    & 25.6 & 4.1  & 871  & 79.0 & 94.4 \\
    RegNetY-4G ~\cite{radosavovic2020designing}     & 21 & 4.0  & 1157  & 80.0 & - \\
    RegNetY-8G ~\cite{radosavovic2020designing}     & 39 & 8.0  & 592  & 81.7 & - \\
    \midrule
    \multicolumn{6}{c}{Transformer}                   \\
    \midrule
    MPViT-T ~\cite{lee2022mpvit} & 5.8 & 1.6 & - & 78.2 & - \\
    \textbf{CageViT-T (Ours)} & 9.91 (14.01) & 1.2 & 1341 & 78.4 & 94.1 \\
    PVT-T ~\cite{wang2021pyramid} & 13.2 & 1.9 & 1038 & 75.1 & 92.4 \\
    \textbf{CageViT-S (Ours)} & 13.53 (17.63) & 1.9 & 1052 & 80.4 & 94.9 \\
    \midrule
    DeiT-S ~\cite{touvron2021training} & 22.1 & 4.6 & 940 & 79.8 & 94.9 \\
    PVT-S ~\cite{wang2021pyramid} & 24.5 & 3.7 & 820 & 79.8 & 94.9 \\
    Swin-T ~\cite{liu2021swin} & 28.29 & 4.5 & 755 & 81.3 & 95.5 \\
    Twins-SVT-S ~\cite{chu2021twins} & 24 & 2.9 & 1059 & 81.7 & - \\
    \textbf{CageViT-B (Ours)} & 24.34 (28.44) & 3.7 & 704 & 82.0 & 95.6 \\
    Focal-Tiny ~\cite{NEURIPS2021_fc1a3682} & 29.1 & 4.9 & - & 82.2 & - \\
    \midrule
    PVT-M ~\cite{wang2021pyramid} & 44.2 & 6.4 & 526 & 81.2 & 95.6 \\
    DeiT-B ~\cite{touvron2021training} & 86.6 & 17.6 & 292 & 81.8 & 95.6 \\
    MViT-B-16 ~\cite{fan2021multiscale} & 37.0 & 7.8 & - & 83.0 & - \\
    Twins-SVT-B ~\cite{chu2021twins} & 56 & 8.6 & 469 & 83.2 & - \\
    Swin-S ~\cite{liu2021swin} & 49.61 & 8.7 & 437 & 83.3 & 96.2 \\
    \textbf{CageViT-L (Ours)} & 43.35 (47.45) & 7.5 & 481 & 83.4 & 96.2 \\
    Focal-Small ~\cite{NEURIPS2021_fc1a3682} & 51.1 & 9.1 & - & 83.5 & - \\
    Swin-B ~\cite{liu2021swin} & 87.77 & 15.4 & 278 & 83.5 & 96.5 \\
    NesT-B ~\cite{zhang2022nested} & 68 & - & - & 83.8 & - \\
    MViT-B-24 ~\cite{fan2021multiscale} & 72.7 & 14.7 & - & 84.0 & - \\
    \midrule
    \multicolumn{6}{c}{Hybrid}                   \\
    \midrule
    CvT-13 ~\cite{wu2021cvt}                      & 20 & 6.7 & - & 81.6 & -   \\
    BoTNet-S1-59 ~\cite{srinivas2021bottleneck}                      & 33.5 & 7.3 & - & 81.7 & -   \\
    CvT-21 ~\cite{wu2021cvt}                      & 32 & 10.1 & - & 82.5 & - \\
    BoTNet-S1-110 ~\cite{srinivas2021bottleneck}                      & 54.7 & 10.9 & - & 82.8 & -   \\
    \bottomrule
  \end{tabular}
\vspace{-4mm}
\end{table*}

\begin{wrapfigure}[18]{L}{0.4\textwidth}
\begin{center}
\includegraphics[width=0.9\linewidth]{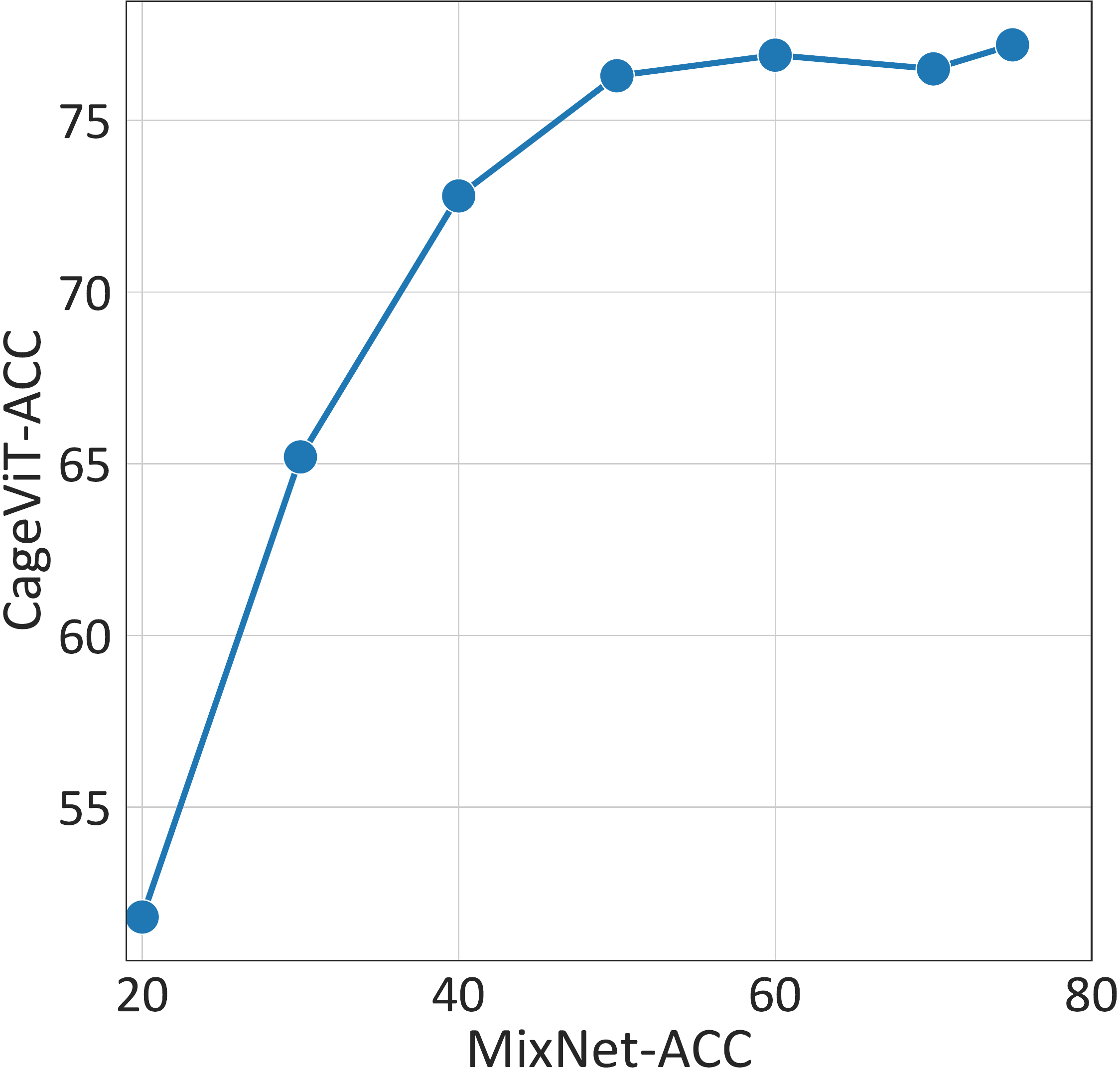}
\end{center}
  \caption{Impact of convolutional model $M_{conv}$ accuracy (Top-1 accuracy) on CageViT-Tiny performance (Top-1 accuracy) on ImageNet-1k dataset.}
\label{fig: impact of M acc}
\end{wrapfigure}

\section{Experiments}

In this section, we conduct experiments on ImageNet-1k for classification task. In the following subsections, we first compared the proposed CageViT with the previous state-of-the-arts on image classification task. Then we adopt ablation studies to validate the core design elements of CageViT.

\subsection{Image Classification on ImageNet-1k}
\label{image classification}

\textbf{Experimental settings.} We evaluate the proposed CageViT model on ImageNet-1k dataset for image classification. The dataset contains 1.28 million training images and 50,000 validation images from 1,000 classes. Most of the experimental setup follows ~\cite{touvron2021training}. We use AdamW optimizer ~\cite{loshchilov2017decoupled} for 300 epochs with cosine decay learning rate scheduler and 5 epochs of linear warm-up. We use batch size of 2048 distributed over 8 GPUs with 256 images per GPU. The initial learning rate is set to 5e-4, weight decay is 0.05 and gradient clipping with a max norm of 5 is applied.

We employ most of the augmentation and regularization strategies from~\cite{touvron2021training}, including RandAugment~\cite{cubuk2020randaugment}, Mixup~\cite{zhang2017mixup}, Cutmix~\cite{yun2019cutmix}, Random erasing~\cite{zhong2020random}, and stochastic depth~\cite{huang2016deep}. The degree of stochastic depth augmentation increases with the size of the model, \textit{i.e.}, 0.1, 0.1, 0.2, 0.3 for CageViT-Tiny, CageViT-Small, CageViT-Base, and CageViT-Large, respectively. During testing on the validation set, the shorter side of the input image is first resized to 256, and a center crop of $224 \times 224$ is used for evaluation. All images are patchified into 256 patches, each containing $14 \times 14$ pixels. All experiments are carried out using the PyTorch~\cite{NEURIPS2019_9015} framework.

\textbf{Results.} Tab.~\ref{table: main} presents comparisons to other backbones, including both Transformer-based ones
and ConvNet-based ones. We can see that, compared to the previous state-of-the-art Transformer-based architectures with similar model complexity, the proposed CageViT achieves significant improvement. For example, for smaller models, CageViT noticeably surpasses the counterpart PVT\cite{wang2021pyramid} architectures with similar complexities: +5.3\% for CageViT-S (80.4\%) over PVT-T (75.1\%). For larger models, CageViT also significantly outperforms the counterpart Swin\cite{liu2021swin} and Twins\cite{chu2021twins} architectures with similar complexities: +0.3\% for CageViT-B (82.0\%) over Twins-SVT-S (81.7\%), and +0.1\% for CageViT-L(83.4\%) over Swin-S(83.3\%) using $224 \times 224$ input. Note that although the Top-1 accuracy of CageViT-B and CageViT-L is slightly lower than that of the same level Focal Transformer\cite{NEURIPS2021_fc1a3682}, this is reasonable given that CageViT has fewer parameters and requires less computational resources.

\subsection{The impact of $M_{conv}$ accuracy}
\label{impact of conv}

\textbf{Experimental settings.} To investigate the impact of $M_{conv}$ accuracy on CageViT performance, we obtained a series of $M_{conv}^i$ (Top-1 $Acc = 20, 30, 40, 50, 60, 70, 75$), models with different Top-1 accuracies using early stopping. For each $M_{conv}^i$, we employ the same experimental setting reported in Sec.~\ref{image classification} for CageViT-Tiny on ImageNet-1k dataset.

\textbf{Results.} Fig.~\ref{fig: impact of M acc} presents the empirical results on the impact of different $M_{conv}^i$. We can see, that after improving the Top-1 accuracy of $M_{conv}$ above 50\%, the accuracy of CageViT-Tiny generally converges to a constant value. Therefore, we can leverage a $M_{conv}$ that is not fully trained into a high accuracy, which means we were able to save a lot of computational costs during the training of $M_{conv}$. Since we would take a weighted average of top-$K$ class activations in CAGA (Sec.~\ref{CAGA}), there is a high probability for the true class label to appear in the top-$K$ labels, despite the moderate accuracy of $M_{conv}$.

\begin{table}[h]
\vspace{-2mm}
\small
\centering
  \caption{The impact of the value $\rho$ on the performance of CageViT. Most of the hyper-parameters are the same as the experimental setting for CageViT-T reported in Sec.~\ref{image classification}, except for the $\rho$ value, resulting in the ratio of selected major tokens. The highest results for Top-1 and Top-5 are shown in \textbf{bold}. The second highest results are marked with \underline{underline}.}
    \begin{tabular}{cccccccccccc}
    \toprule
    & 0 & 10 & 20 & 30 & 40 & 50 & 60 & 70 & 80 & 90 & 100 \\
    \midrule
    Top-1 (\%) & 29.3 & 62.1 & \textbf{78.4} & \underline{77.5} & 76.9 & \textbf{78.4} & 76.1 & \underline{77.5} & 76.0 & 72.9 & 75.3 \\
    Top-5 (\%) & 51.9 & 79.9 & \textbf{94.1} & 92.5 & 92.2 & \underline{93.9} & 92.5 & 93.1 & 92.3 & 91.8 & 92.4 \\
    \#Params (M) & 0.51 & 4.7 & 9.5 & 11.3 & 15.8 & 21.2 & 27.1 & 32.7 & 38.9 & 45.1 & 54.3 \\
    \bottomrule
    \end{tabular}
\label{table: rho impact}
\end{table}

\begin{table}
\vspace{-2mm}
\small
  \caption{The impact of the value $K$ on the performance of CageViT. Most of the hyper-parameters are the same as the experimental setting for CageViT-Tiny reported in Sec.~\ref{image classification}, except for the $K$ value, resulting in the selected top-$K$ labels while calculating the weighted average Grad-CAM++. The best result on Top-1 and Top-5 are in \textbf{bold}.}
  \centering
  \begin{tabular}{lccccccccccccc}
    \toprule
    $K$   &  1  &   2   &   3   &  4  &   5   &   6   &  7  &   8   &   9   &  10  &   20   &   50   &  100 \\
    \midrule
Top-1 (\%)              & 61.4                  & 62.2                  & 69.4                  & 68.3                  & 71.6                  & 74.1                  & 71.4                  & 71.9                  & \textbf{78.4} & 76.3                   & 72.8                   & 72.1                   & 69.4                    \\
Top-5 (\%)              & 78.8                  & 75.6                  & 78.8                  & 83.8                  & 90.2                  & 91.3                  & 89.5                  & 90.0                  & \textbf{94.1}                           & 90.7                   & 89.8                   & 89.4                   & 81.6                    \\
    \bottomrule
  \end{tabular}
\label{table: K impact}
\end{table}

\subsection{The impact of different $\rho$ values}
\label{impact of rho}

\textbf{Experimental settings.} We test the effect of different $\rho$ values on CageViT-Tiny performance. Most of the hyperparameters are the same as the experimental setting for CageViT-Tiny reported in Sec.~\ref{image classification}, except for the value of $\rho$, which determines the ratio of selected major tokens.

\textbf{Results.} Tab.~\ref{table: rho impact} presents the empirical results on the impact of different $\rho$. We can see, that the CageViT-Tiny performs best at $\rho$ values of 20 and 50, with a relatively small number of parameters (9.5 and 21.2, respectively).

\subsection{The impact of different $K$ values}
\label{impact of K}

\textbf{Experimental settings.} To investigate the impact of the different number of selected top-$K$ labels during calculating weighted average Grad-CAM++, \textit{i.e.} $K$ values, on CageViT-Tiny performance, we conduct experiments on different $K$ values. Most of the hyper-parameters are the same as the experimental setting for CageViT-Tiny reported in Sec.~\ref{image classification}, except for the $K$ value.

\textbf{Results.} Tab.~\ref{table: K impact} presents the empirical results on the impact of different $k$ values. We can see, that the CageViT-Tiny performs relatively worse with $K=1,2$, which means that the $M_{conv}$ cannot precisely classify the correct labels. As we increase the $K$ value, the model performs much better when $K$ is around 10. However, the accuracy gets lower with $K>10$. This means that an increase in the value of K may make the major tokens come more from small items in the background, which disrupts the model's judgment of fine-grained information.

\begin{table*}[!ht]
\small
\vspace{-4mm}
  \caption{Ablation of CageViT-Tiny: We report Top-1, Top-5 accuracy and number of parameters for different model designs on $224 \times 224$ image resolution. Our results indicate that the use of CAGA (as described in Section~\ref{CAGA}) significantly reduces computational cost, while the specifically designed Gated Linear SRA improves performance compared to the original Linear SRA.}
  \centering
  \begin{tabular}{lccc}
    \toprule
    Model   &  Top-1 (\%) & Top-5 (\%) & \#Params (M)   \\
    \midrule
    ViT  & 77.91 & 93.83 & 64.8 \\
    ViT + CAGA  & 72.58 & 90.31 & 19.2 \\
    ViT + Linear SRA  & 77.94 & 93.90 & 25.5 \\
    ViT + CAGA + Linear SRA  & 75.40 & 91.53 & 9.16 \\
    ViT + CAGA + Gated Linear SRA  & 78.38 & 94.12 & 9.91 \\
    \bottomrule
  \end{tabular}
\label{table: design ablation}
\end{table*}

\begin{table*}[!ht]
\small
\vspace{-4mm}
  \caption{Ablations of CageViT-Tiny on different class activation methods. We report the Top-1 accuracy, Top-5 accuracy, and class activation method throughput. All the models are trained and evaluated on $224 \times 224$ resolution. The best records are marked in \textbf{bold}.}
  \centering
  \begin{tabular}{lccc}
    \toprule
    CAM   &  Top-1 (\%) & Top-5 (\%) & Throughput   \\
    \midrule
    Grad-CAM~\cite{selvaraju2017grad}  & 74.50 & 92.17 & \textbf{8492} \\
    Grad-CAM++~\cite{chattopadhay2018grad}  & \textbf{78.38} & \textbf{94.12} & 5813 \\
    Score-CAM~\cite{wang2020score}   & 73.81 & 91.84 & 6924 \\
    Ablation-CAM~\cite{ramaswamy2020ablation}  & 77.19 & 93.35 & 2750 \\
    Eigen-CAM~\cite{muhammad2020eigen}  & 76.43 & 92.90 & 4815 \\
    \midrule
    Grad-CAM++ w/o confidence  & 75.42 & 92.58 & 6245 \\
    \midrule
    Random Selection  & 47.53 & 61.40 & N/A \\
    \bottomrule
  \end{tabular}
\label{table: CAM ablation}
\end{table*}

\subsection{Ablation Study}

In this section, we report the ablation studies of the proposed CageViT, using ImageNet-1k image classification. All experimental setups are consistent with those reported in Sec.~\ref{image classification}, except for the different modules (\textit{e.g.} CAGA, Linear SRA) equipped on the model.

\subsubsection{Ablation on the proposed designs}

From Tab.~\ref{table: design ablation}, we see that all three designs can improve the model in terms of performance, parameter number, or computation overhead.

\textbf{CAGA can vastly reduce the computational cost.} After adding the CAGA module into ViT, the number of parameters is reduced from 64.8M to 18.4M, while losing some accuracy. 

\textbf{Gated Linear SRA is more suitable than Linear SRA while the model is equipped with CAGA.}. The top-1 accuracy is improved from 75.40\% to 78.38\% after replacing Linear SRA with Gated Linear SRA, while only increasing 0.75M parameters.

\subsubsection{Ablation on different class activation method}

In Tab.~\ref{table: CAM ablation}, we present the performance of various class activation methods for selecting important tokens. These include Grad-CAM~\cite{selvaraju2017grad}, Grad-CAM++~\cite{chattopadhay2018grad}, Score-CAM~\cite{wang2020score}, Ablation-CAM~\cite{ramaswamy2020ablation}, and Eigen-CAM~\cite{muhammad2020eigen}. All the implementation of class activation methods comes from PyTroch-Grad-CAM~\cite{jacobgilpytorchcam}. Among the methods evaluated, Grad-CAM++ achieved the highest Top-1 accuracy (78.38\%) and Top-5 accuracy (94.12\%), while also being computationally efficient (5813 throughput).

We also report the results of removing the influence of confidence when using Grad-CAM++, \textit{i.e.} setting all the $z_k$ to be 1 in the Eq.~(\ref{equation weighted}). The top-1 accuracy is reduced from 78.38\% to 75.40\%, which means that the quality of the selected major tokens gets lower, demonstrating the effectiveness of weighted average Grad-CAM++ depicted in Sec.~\ref{CAGA}.

As a point of comparison, we also report the results of randomly selecting $\rho\%$ of the tokens as a lower bound for token selection methods. It is evident that the performance of randomly selected tokens is significantly lower than that of using CAM methods.

%% file: 5_conclusion.tex
\section{Conclusion}

In this paper, we introduce a novel, efficient transformer model called CageViT. We propose a two-fold approach to accelerate the performance of transformers. Specifically, we use convolutional activation to preprocess the token after patchifying the image to select and rearrange major and minor tokens, which greatly reduces computation cost through the use of an additional fusion layer. Instead of directly using the class activation map of the convolutional model, we design a new weighted class activation to reduce the requirements for model quality of the additionally introduced convolutional model. Additionally, we propose Gated Linear SRA to bring the fusion tokens deeper into the core calculation of the attention mechanism, facilitating communication between major and fusion tokens. We evaluate CageViT on the image classification task and conduct an ablation study to validate the design of various modules. Corresponding results demonstrate the effectiveness of CageViT.

%% file: main.bbl
\begin{thebibliography}{10}

\bibitem{bahdanau2014neural}
Dzmitry Bahdanau, Kyunghyun Cho, and Yoshua Bengio.
\newblock Neural machine translation by jointly learning to align and
  translate.
\newblock {\em arXiv preprint arXiv:1409.0473}, 2014.

\bibitem{beltagy2020longformer}
Iz~Beltagy, Matthew~E Peters, and Arman Cohan.
\newblock Longformer: The long-document transformer.
\newblock {\em arXiv preprint arXiv:2004.05150}, 2020.

\bibitem{carion2020end}
Nicolas Carion, Francisco Massa, Gabriel Synnaeve, Nicolas Usunier, Alexander
  Kirillov, and Sergey Zagoruyko.
\newblock End-to-end object detection with transformers.
\newblock In {\em European Conference on Computer Vision}, pages 213--229.
  Springer, 2020.

\bibitem{chattopadhay2018grad}
Aditya Chattopadhay, Anirban Sarkar, Prantik Howlader, and Vineeth~N
  Balasubramanian.
\newblock Grad-cam++: Generalized gradient-based visual explanations for deep
  convolutional networks.
\newblock In {\em 2018 IEEE winter conference on applications of computer
  vision (WACV)}, pages 839--847. IEEE, 2018.

\bibitem{choromanski2020rethinking}
Krzysztof Choromanski, Valerii Likhosherstov, David Dohan, Xingyou Song,
  Andreea Gane, Tamas Sarlos, Peter Hawkins, Jared Davis, Afroz Mohiuddin,
  Lukasz Kaiser, et~al.
\newblock Rethinking attention with performers.
\newblock {\em arXiv preprint arXiv:2009.14794}, 2020.

\bibitem{chu2021twins}
Xiangxiang Chu, Zhi Tian, Yuqing Wang, Bo~Zhang, Haibing Ren, Xiaolin Wei,
  Huaxia Xia, and Chunhua Shen.
\newblock Twins: Revisiting the design of spatial attention in vision
  transformers.
\newblock {\em Advances in Neural Information Processing Systems},
  34:9355--9366, 2021.

\bibitem{cubuk2020randaugment}
Ekin~D Cubuk, Barret Zoph, Jonathon Shlens, and Quoc~V Le.
\newblock Randaugment: Practical automated data augmentation with a reduced
  search space.
\newblock In {\em Proceedings of the IEEE/CVF Conference on Computer Vision and
  Pattern Recognition Workshops}, pages 702--703, 2020.

\bibitem{dai2019transformer}
Zihang Dai, Zhilin Yang, Yiming Yang, Jaime Carbonell, Quoc~V Le, and Ruslan
  Salakhutdinov.
\newblock Transformer-xl: Attentive language models beyond a fixed-length
  context.
\newblock {\em arXiv preprint arXiv:1901.02860}, 2019.

\bibitem{devlin2018bert}
Jacob Devlin, Ming-Wei Chang, Kenton Lee, and Kristina Toutanova.
\newblock Bert: Pre-training of deep bidirectional transformers for language
  understanding.
\newblock {\em arXiv preprint arXiv:1810.04805}, 2018.

\bibitem{dosovitskiy2020image}
Alexey Dosovitskiy, Lucas Beyer, Alexander Kolesnikov, Dirk Weissenborn,
  Xiaohua Zhai, Thomas Unterthiner, Mostafa Dehghani, Matthias Minderer, Georg
  Heigold, Sylvain Gelly, et~al.
\newblock An image is worth 16x16 words: Transformers for image recognition at
  scale.
\newblock {\em arXiv preprint arXiv:2010.11929}, 2020.

\bibitem{fan2021multiscale}
Haoqi Fan, Bo~Xiong, Karttikeya Mangalam, Yanghao Li, Zhicheng Yan, Jitendra
  Malik, and Christoph Feichtenhofer.
\newblock Multiscale vision transformers.
\newblock In {\em Proceedings of the IEEE/CVF International Conference on
  Computer Vision}, pages 6824--6835, 2021.

\bibitem{jacobgilpytorchcam}
Jacob Gildenblat and contributors.
\newblock Pytorch library for cam methods.
\newblock \url{https://github.com/jacobgil/pytorch-grad-cam}, 2021.

\bibitem{he2016deep}
Kaiming He, Xiangyu Zhang, Shaoqing Ren, and Jian Sun.
\newblock Deep residual learning for image recognition.
\newblock In {\em Proceedings of the IEEE conference on computer vision and
  pattern recognition}, pages 770--778, 2016.

\bibitem{huang2016deep}
Gao Huang, Yu~Sun, Zhuang Liu, Daniel Sedra, and Kilian~Q Weinberger.
\newblock Deep networks with stochastic depth.
\newblock In {\em European conference on computer vision}, pages 646--661.
  Springer, 2016.

\bibitem{kitaev2020reformer}
Nikita Kitaev, {\L}ukasz Kaiser, and Anselm Levskaya.
\newblock Reformer: The efficient transformer.
\newblock {\em arXiv preprint arXiv:2001.04451}, 2020.

\bibitem{lee2022mpvit}
Youngwan Lee, Jonghee Kim, Jeffrey Willette, and Sung~Ju Hwang.
\newblock Mpvit: Multi-path vision transformer for dense prediction.
\newblock In {\em Proceedings of the IEEE/CVF Conference on Computer Vision and
  Pattern Recognition}, pages 7287--7296, 2022.

\bibitem{lin2021end}
Kevin Lin, Lijuan Wang, and Zicheng Liu.
\newblock End-to-end human pose and mesh reconstruction with transformers.
\newblock In {\em Proceedings of the IEEE/CVF Conference on Computer Vision and
  Pattern Recognition}, pages 1954--1963, 2021.

\bibitem{liu2021swin}
Ze~Liu, Yutong Lin, Yue Cao, Han Hu, Yixuan Wei, Zheng Zhang, Stephen Lin, and
  Baining Guo.
\newblock Swin transformer: Hierarchical vision transformer using shifted
  windows.
\newblock In {\em Proceedings of the IEEE/CVF International Conference on
  Computer Vision}, pages 10012--10022, 2021.

\bibitem{loshchilov2017decoupled}
Ilya Loshchilov and Frank Hutter.
\newblock Decoupled weight decay regularization.
\newblock {\em arXiv preprint arXiv:1711.05101}, 2017.

\bibitem{muhammad2020eigen}
Mohammed~Bany Muhammad and Mohammed Yeasin.
\newblock Eigen-cam: Class activation map using principal components.
\newblock In {\em 2020 International Joint Conference on Neural Networks
  (IJCNN)}, pages 1--7. IEEE, 2020.

\bibitem{parikh2016decomposable}
Ankur~P Parikh, Oscar T{\"a}ckstr{\"o}m, Dipanjan Das, and Jakob Uszkoreit.
\newblock A decomposable attention model for natural language inference.
\newblock {\em arXiv preprint arXiv:1606.01933}, 2016.

\bibitem{parmar2018image}
Niki Parmar, Ashish Vaswani, Jakob Uszkoreit, Lukasz Kaiser, Noam Shazeer,
  Alexander Ku, and Dustin Tran.
\newblock Image transformer.
\newblock In {\em International Conference on Machine Learning}, pages
  4055--4064. PMLR, 2018.

\bibitem{NEURIPS2019_9015}
Adam Paszke, Sam Gross, Francisco Massa, Adam Lerer, James Bradbury, Gregory
  Chanan, Trevor Killeen, Zeming Lin, Natalia Gimelshein, Luca Antiga, Alban
  Desmaison, Andreas Kopf, Edward Yang, Zachary DeVito, Martin Raison, Alykhan
  Tejani, Sasank Chilamkurthy, Benoit Steiner, Lu~Fang, Junjie Bai, and Soumith
  Chintala.
\newblock Pytorch: An imperative style, high-performance deep learning library.
\newblock In H.~Wallach, H.~Larochelle, A.~Beygelzimer, F.~d\textquotesingle
  Alch\'{e}-Buc, E.~Fox, and R.~Garnett, editors, {\em Advances in Neural
  Information Processing Systems 32}, pages 8024--8035. Curran Associates,
  Inc., 2019.

\bibitem{qiu2019blockwise}
Jiezhong Qiu, Hao Ma, Omer Levy, Scott Wen-tau Yih, Sinong Wang, and Jie Tang.
\newblock Blockwise self-attention for long document understanding.
\newblock {\em arXiv preprint arXiv:1911.02972}, 2019.

\bibitem{radford2019language}
Alec Radford, Jeff Wu, Rewon Child, David Luan, Dario Amodei, and Ilya
  Sutskever.
\newblock Language models are unsupervised multitask learners.
\newblock 2019.

\bibitem{radosavovic2020designing}
Ilija Radosavovic, Raj~Prateek Kosaraju, Ross Girshick, Kaiming He, and Piotr
  Doll{\'a}r.
\newblock Designing network design spaces.
\newblock In {\em Proceedings of the IEEE/CVF Conference on Computer Vision and
  Pattern Recognition}, pages 10428--10436, 2020.

\bibitem{ramaswamy2020ablation}
Harish~Guruprasad Ramaswamy et~al.
\newblock Ablation-cam: Visual explanations for deep convolutional network via
  gradient-free localization.
\newblock In {\em Proceedings of the IEEE/CVF Winter Conference on Applications
  of Computer Vision}, pages 983--991, 2020.

\bibitem{selvaraju2017grad}
Ramprasaath~R Selvaraju, Michael Cogswell, Abhishek Das, Ramakrishna Vedantam,
  Devi Parikh, and Dhruv Batra.
\newblock Grad-cam: Visual explanations from deep networks via gradient-based
  localization.
\newblock In {\em Proceedings of the IEEE international conference on computer
  vision}, pages 618--626, 2017.

\bibitem{shoeybi2019megatron}
Mohammad Shoeybi, Mostofa Patwary, Raul Puri, Patrick LeGresley, Jared Casper,
  and Bryan Catanzaro.
\newblock Megatron-lm: Training multi-billion parameter language models using
  model parallelism.
\newblock {\em arXiv preprint arXiv:1909.08053}, 2019.

\bibitem{srinivas2021bottleneck}
Aravind Srinivas, Tsung-Yi Lin, Niki Parmar, Jonathon Shlens, Pieter Abbeel,
  and Ashish Vaswani.
\newblock Bottleneck transformers for visual recognition.
\newblock In {\em Proceedings of the IEEE/CVF conference on computer vision and
  pattern recognition}, pages 16519--16529, 2021.

\bibitem{tan2019mixconv}
Mingxing Tan and Quoc~V Le.
\newblock Mixconv: Mixed depthwise convolutional kernels.
\newblock {\em arXiv preprint arXiv:1907.09595}, 2019.

\bibitem{tay2020synthesizer}
Yi~Tay, Dara Bahri, Donald Metzler, Da-Cheng Juan, Zhe Zhao, and Che Zheng.
\newblock Synthesizer: Rethinking self-attention in transformer models.
\newblock {\em icml}, 2020.

\bibitem{tay2020sparse}
Yi~Tay, Dara Bahri, Liu Yang, Donald Metzler, and Da-Cheng Juan.
\newblock Sparse sinkhorn attention.
\newblock In {\em International Conference on Machine Learning}, pages
  9438--9447. PMLR, 2020.

\bibitem{tay2020long}
Yi~Tay, Mostafa Dehghani, Samira Abnar, Yikang Shen, Dara Bahri, Philip Pham,
  Jinfeng Rao, Liu Yang, Sebastian Ruder, and Donald Metzler.
\newblock Long range arena: A benchmark for efficient transformers.
\newblock {\em arXiv preprint arXiv:2011.04006}, 2020.

\bibitem{tay2020efficient}
Yi~Tay, Mostafa Dehghani, Dara Bahri, and Donald Metzler.
\newblock Efficient transformers: A survey.
\newblock {\em arXiv preprint arXiv:2009.06732}, 2020.

\bibitem{touvron2021training}
Hugo Touvron, Matthieu Cord, Matthijs Douze, Francisco Massa, Alexandre
  Sablayrolles, and Herv{\'e} J{\'e}gou.
\newblock Training data-efficient image transformers \& distillation through
  attention.
\newblock In {\em International Conference on Machine Learning}, pages
  10347--10357. PMLR, 2021.

\bibitem{vaswani2017attention}
Ashish Vaswani, Noam Shazeer, Niki Parmar, Jakob Uszkoreit, Llion Jones,
  Aidan~N Gomez, {\L}ukasz Kaiser, and Illia Polosukhin.
\newblock Attention is all you need.
\newblock In {\em Advances in neural information processing systems}, pages
  5998--6008, 2017.

\bibitem{wang2020score}
Haofan Wang, Zifan Wang, Mengnan Du, Fan Yang, Zijian Zhang, Sirui Ding, Piotr
  Mardziel, and Xia Hu.
\newblock Score-cam: Score-weighted visual explanations for convolutional
  neural networks.
\newblock In {\em Proceedings of the IEEE/CVF conference on computer vision and
  pattern recognition workshops}, pages 24--25, 2020.

\bibitem{wang2020linformer}
Sinong Wang, Belinda~Z Li, Madian Khabsa, Han Fang, and Hao Ma.
\newblock Linformer: Self-attention with linear complexity.
\newblock {\em arXiv preprint arXiv:2006.04768}, 2020.

\bibitem{wang2021pyramid}
Wenhai Wang, Enze Xie, Xiang Li, Deng-Ping Fan, Kaitao Song, Ding Liang, Tong
  Lu, Ping Luo, and Ling Shao.
\newblock Pyramid vision transformer: A versatile backbone for dense prediction
  without convolutions.
\newblock In {\em Proceedings of the IEEE/CVF International Conference on
  Computer Vision}, pages 568--578, 2021.

\bibitem{wang2022pvt}
Wenhai Wang, Enze Xie, Xiang Li, Deng-Ping Fan, Kaitao Song, Ding Liang, Tong
  Lu, Ping Luo, and Ling Shao.
\newblock Pvt v2: Improved baselines with pyramid vision transformer.
\newblock {\em Computational Visual Media}, pages 1--10, 2022.

\bibitem{wu2021cvt}
Haiping Wu, Bin Xiao, Noel Codella, Mengchen Liu, Xiyang Dai, Lu~Yuan, and Lei
  Zhang.
\newblock Cvt: Introducing convolutions to vision transformers.
\newblock In {\em Proceedings of the IEEE/CVF International Conference on
  Computer Vision}, pages 22--31, 2021.

\bibitem{NEURIPS2021_fc1a3682}
Jianwei Yang, Chunyuan Li, Pengchuan Zhang, Xiyang Dai, Bin Xiao, Lu~Yuan, and
  Jianfeng Gao.
\newblock Focal attention for long-range interactions in vision transformers.
\newblock In M.~Ranzato, A.~Beygelzimer, Y.~Dauphin, P.S. Liang, and J.~Wortman
  Vaughan, editors, {\em Advances in Neural Information Processing Systems},
  volume~34, pages 30008--30022. Curran Associates, Inc., 2021.

\bibitem{yang2019xlnet}
Zhilin Yang, Zihang Dai, Yiming Yang, Jaime Carbonell, Russ~R Salakhutdinov,
  and Quoc~V Le.
\newblock Xlnet: Generalized autoregressive pretraining for language
  understanding.
\newblock {\em Advances in neural information processing systems}, 32, 2019.

\bibitem{yun2019cutmix}
Sangdoo Yun, Dongyoon Han, Seong~Joon Oh, Sanghyuk Chun, Junsuk Choe, and
  Youngjoon Yoo.
\newblock Cutmix: Regularization strategy to train strong classifiers with
  localizable features.
\newblock In {\em Proceedings of the IEEE/CVF international conference on
  computer vision}, pages 6023--6032, 2019.

\bibitem{zaheer2020big}
Manzil Zaheer, Guru Guruganesh, Kumar~Avinava Dubey, Joshua Ainslie, Chris
  Alberti, Santiago Ontanon, Philip Pham, Anirudh Ravula, Qifan Wang, Li~Yang,
  et~al.
\newblock Big bird: Transformers for longer sequences.
\newblock In {\em NeurIPS}, 2020.

\bibitem{zhang2017mixup}
Hongyi Zhang, Moustapha Cisse, Yann~N Dauphin, and David Lopez-Paz.
\newblock mixup: Beyond empirical risk minimization.
\newblock {\em arXiv preprint arXiv:1710.09412}, 2017.

\bibitem{zhang2021multi}
Pengchuan Zhang, Xiyang Dai, Jianwei Yang, Bin Xiao, Lu~Yuan, Lei Zhang, and
  Jianfeng Gao.
\newblock Multi-scale vision longformer: A new vision transformer for
  high-resolution image encoding.
\newblock In {\em Proceedings of the IEEE/CVF International Conference on
  Computer Vision}, pages 2998--3008, 2021.

\bibitem{zhang2022nested}
Zizhao Zhang, Han Zhang, Long Zhao, Ting Chen, Sercan~{\"O} Arik, and Tomas
  Pfister.
\newblock Nested hierarchical transformer: Towards accurate, data-efficient and
  interpretable visual understanding.
\newblock In {\em Proceedings of the AAAI Conference on Artificial
  Intelligence}, volume~36, pages 3417--3425, 2022.

\bibitem{zheng2021rethinking}
Sixiao Zheng, Jiachen Lu, Hengshuang Zhao, Xiatian Zhu, Zekun Luo, Yabiao Wang,
  Yanwei Fu, Jianfeng Feng, Tao Xiang, Philip~HS Torr, et~al.
\newblock Rethinking semantic segmentation from a sequence-to-sequence
  perspective with transformers.
\newblock In {\em Proceedings of the IEEE/CVF Conference on Computer Vision and
  Pattern Recognition}, pages 6881--6890, 2021.

\bibitem{zhong2020random}
Zhun Zhong, Liang Zheng, Guoliang Kang, Shaozi Li, and Yi~Yang.
\newblock Random erasing data augmentation.
\newblock In {\em Proceedings of the AAAI conference on artificial
  intelligence}, volume~34, pages 13001--13008, 2020.

\bibitem{zhou2018end}
Luowei Zhou, Yingbo Zhou, Jason~J Corso, Richard Socher, and Caiming Xiong.
\newblock End-to-end dense video captioning with masked transformer.
\newblock In {\em Proceedings of the IEEE Conference on Computer Vision and
  Pattern Recognition}, pages 8739--8748, 2018.

\end{thebibliography}
